\documentclass[11pt]{article}
\usepackage[preprint]{acl}
\usepackage{times}
\usepackage{latexsym}
\usepackage{amsmath}
\usepackage{amssymb}
\usepackage{booktabs}
\usepackage{makecell}
\usepackage{pifont}
\usepackage[T1]{fontenc}
\usepackage[utf8]{inputenc}
\usepackage{microtype}
\usepackage{tabularx}
\usepackage{booktabs}
\usepackage{inconsolata}
\usepackage{comment}
\usepackage{graphicx}
\usepackage{booktabs}
\usepackage{pgfplots}
\pgfplotsset{compat=1.18}

\usepackage{hyperref}
\usepackage{url}
\usepackage{microtype}
\usepackage{tabularx}
\usepackage{booktabs}
\usepackage{inconsolata}
\usepackage{booktabs}
\usepackage{threeparttable}
\usepackage{listings}
\usepackage{float}

\lstdefinestyle{judgeprompt}{
    basicstyle=\ttfamily\footnotesize,
    breaklines=true,
    breakatwhitespace=false,
    columns=fullflexible,
    keepspaces=true,
    showstringspaces=false,
    frame=single,
    framerule=0.4pt,
    framesep=5pt,
    xleftmargin=6pt,
    xrightmargin=6pt,
    aboveskip=8pt,
    belowskip=8pt
}

\title{IntentDelta: A Human-Constructed Benchmark for Evolving User Intent}
\author{
    \textbf{Jieyu Zhou}\textsuperscript{1},
    \textbf{Zijian Wang}\textsuperscript{1},
    \textbf{Yun Wang}\textsuperscript{2}
    \\
    \\
    \textsuperscript{1}Georgia Institute of Technology
    \quad
    \textsuperscript{2}Microsoft Research
    \\
    \textsuperscript{1}Atlanta, USA
    \quad
    \textsuperscript{2}Hong Kong SAR, China
    \\
    \texttt{\{jzhou625, zwang3613\}@gatech.edu}
    \\
    \texttt{wangyun@microsoft.com}
}
\begin{document}

\maketitle

\begin{abstract}
Language-model agents must adapt as users revise their requirements and priorities during complex tasks. We introduce IntentDelta, a human-constructed multi-turn benchmark with 638 trajectories and 3,766 turns across WebShop and TravelPlanner. Each turn pairs a user utterance with a cumulative intention state and human reference actions. The trajectories are cross-verified and cover feasibility- and preference-driven changes in constraints and priorities. We evaluate eleven agents on intention recovery and action quality. Average Intention Recovery is 33.1\% in TravelPlanner and 44.7\% in WebShop, while average Action Score is 31.6\% and 41.0\%, respectively. Agents struggle to preserve earlier requirements and distinguish soft preferences from must-have constraints. Correct intention recovery also does not guarantee successful execution, highlighting challenges in translating evolving user intentions into actions.
\end{abstract}

\begin{figure*}[t]
    \centering
    \includegraphics[width=\textwidth]{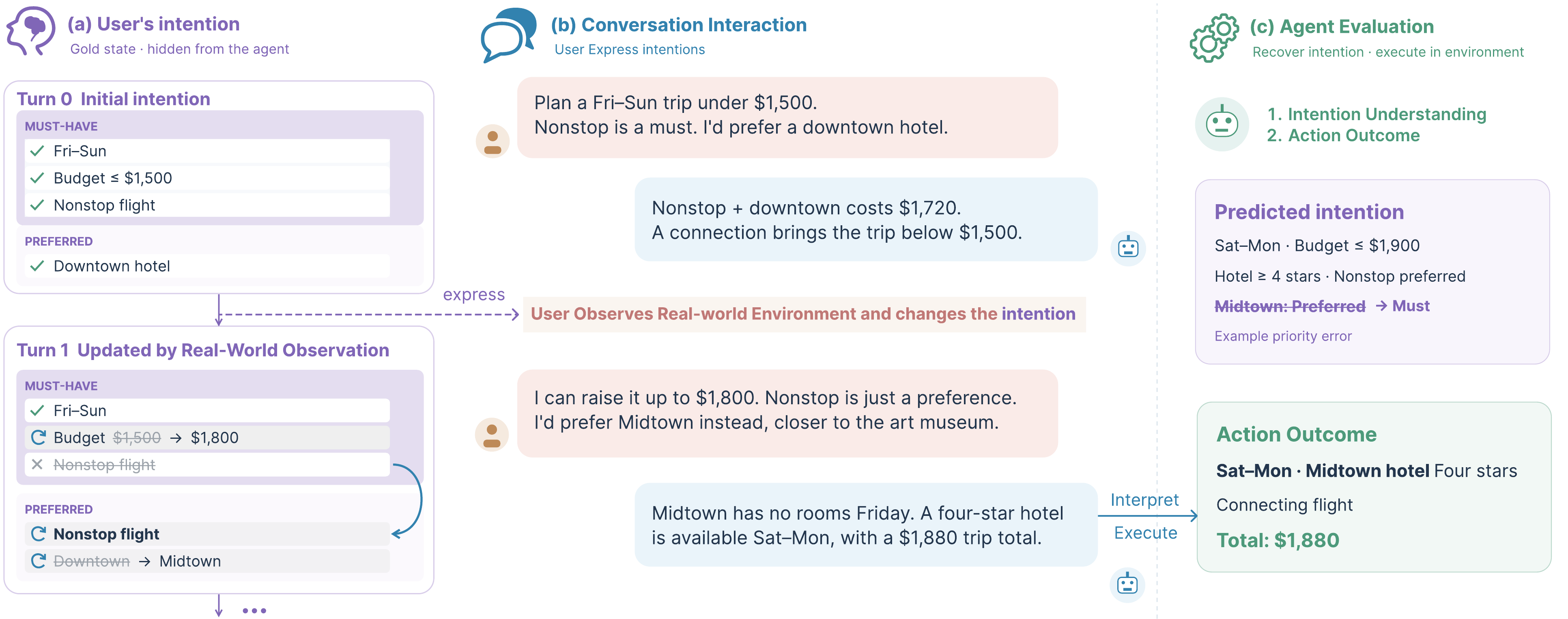}
    \caption{
        Overview of evolving user intentions and agent evaluation.
        \textbf{(a) User's intention:}
        The latent intention, hidden from the agent, evolves in
        response to real-world observations, including infeasibility
        and changes in preference.
        \textbf{(b) Conversation:}
        The user expresses the revised intention through subsequent
        utterances. In this example, the user relaxes the budget,
        demotes the nonstop requirement to a preference, and changes
        the preferred hotel location.
        \textbf{(c) Agent evaluation:}
        The agent is evaluated on both intention understanding
        and action outcomes.
    }
    \label{fig:overview}
\end{figure*}
\section{Introduction}
Language-model agents are increasingly expected to assist users with complex, multi-step tasks such as product search and travel planning \cite{zhou2024webarena, xie2024osworld, ma2024agentboard, xie2024travelplanner, yao2022webshop}. Success in these settings depends not only on executing tools or producing a plausible final answer, but also on maintaining an accurate understanding of what the user currently wants. In practice, user intentions are often not fully specified at the beginning of an interaction. Users may revise their requirements after observing concrete results: a set of products may reveal that the original combination of price and attributes is infeasible, while a proposed itinerary may cause the user to reconsider the trade-off between cost and travel time. Such feedback may add, relax, replace, or re-prioritize a constraint, requiring the agent to preserve requirements that remain valid while discarding those that have been superseded.

\begin{table*}[t]
\centering
\small
\setlength{\tabcolsep}{3pt}
\renewcommand{\arraystretch}{1.12}

\begin{tabular*}{\textwidth}{@{\extracolsep{\fill}}lcccccc@{}}
\toprule[0.7pt]
\textbf{Benchmark} &
\makecell{\textbf{Data}\\\textbf{scale}} &
\makecell{\textbf{Multi-}\\\textbf{domain}} &
\textbf{Construction} &
\makecell{\textbf{Trade-off}\\\textbf{representation}} &
\makecell{\textbf{Change}\\\textbf{conditions}} &
\makecell{\textbf{Human action}\\\textbf{references}} \\
\midrule[0.4pt]

\makecell[l]{Frames\\{\scriptsize\cite{el2017frames}}} &
1,369 dialogues &
\ding{55} &
Human &
--- &
Infeasibility &
\ding{55} \\
\addlinespace[6pt]

\makecell[l]{Flex-TravelPlanner\\{\scriptsize\cite{oh2025flex}}} &
\makecell{120 queries\\+ 134 scenarios} &
\ding{55} &
Synthetic &
Hard/soft &
--- &
NR \\
\addlinespace[6pt]

\makecell[l]{COMPASS\\{\scriptsize\cite{qin2025compass}}} &
281 tasks &
\ding{55} &
LLM-generated &
\makecell{Hard constraints\\+ utility} &
--- &
\ding{55} \\
\addlinespace[6pt]

\makecell[l]{Evolving User Intent\\{\scriptsize\cite{tack2026llms}}} &
\makecell{450 evaluated\\source tasks} &
\ding{51} &
LLM-generated &
--- &
--- &
\ding{55} \\
\addlinespace[3pt]

\midrule[0.4pt]
\textbf{IntentDelta} &
\makecell{638 trajectories\\3,766 turns} &
\ding{51} &
Human &
\makecell{\textbf{Evolving}\\\textbf{priorities}} &
\makecell{Feasibility\\+ preference} &
\ding{51} \\
\bottomrule[0.7pt]
\end{tabular*}

\caption{Benchmark comparison. Multi-domain denotes domain coverage;
human action references denote per-turn human reference executions or
plans. Dashes indicate no explicit priority/change-condition modeling;
NR means not reported. Scale units are retained.}
\label{tab:benchmark_comparison}
\end{table*}

Existing benchmarks study requirement revision and task switching \cite{el2017frames,tack2026llms}, as well as trade-offs between hard constraints and soft preferences \cite{oh2025flex,qin2025compass}. As summarized in Table~\ref{tab:benchmark_comparison}, existing benchmarks have three main limitations. \textit{First}, most rely on synthesized intention updates or simulated user interactions. Synthesis alone provides limited evidence that these changes reflect plausible human decisions. \textit{Second}, existing benchmarks offer limited coverage of users' decision making, including how they weigh competing requirements and under what conditions they revise their intentions. \textit{Third}, human reference actions for each turn's current intention are not consistently provided, limiting comparisons between agent outcomes and human execution as requirements evolve.

To address these gaps, we introduce \textbf{IntentDelta}, a multi-turn benchmark for evaluating whether agents can maintain and act on evolving user intentions. Human annotators construct and cross-verify episodes in WebShop \cite{yao2022webshop} and TravelPlanner \cite{xie2024travelplanner}, pairing natural-language updates with cumulative annotations of active constraints and priorities, as well as human reference actions at each turn. The benchmark covers feasibility- and preference-driven changes, including additions, relaxations, overrides, and reprioritizations. We evaluate intention recovery and action quality separately at each turn, supporting joint analysis of errors in the agent's reported intention state and failures in its resulting actions.

We evaluate eleven LLM agents in WebShop and TravelPlanner using a shared ReAct framework. Our experiments reveal substantial weaknesses: average Intention Recovery is 33.1\% in TravelPlanner and 44.7\% in WebShop, while average Action Score is 31.6\% and 41.0\%, respectively. Failure analysis shows that agents lose previously established requirements, and treat soft preferences as must-have constraints. Higher intention recovery is associated with better action scores, but successful execution remains challenging even when the reported intention is fully recovered. We hope IntentDelta supports research on maintaining evolving user intentions and informs the design of targeted clarification strategies for more reliable task execution.

\section{Related Work}

Existing agent benchmarks typically evaluate task completion under fixed user requirements \citep{yao2022webshop,xie2024travelplanner,liu2024agentbench}. Conversational evaluations further examine multi-turn instruction following \citep{li2025structflowbench}, simulated user interactions \citep{yao2025tau,mehri2026goal}, or clarification of underspecified tasks \citep{kumar2020clarq,li2026questbench}. However, multi-turn interaction alone does not establish whether agents can maintain user requirements as they are added, removed, overridden, or reprioritized in response to task results. IntentDelta preserves grounded task execution while making the user specification a controlled, time-varying component of the episode.

Task-oriented dialogue and conversational recommendation have long studied evolving user requirements. Dialogue state tracking represents needs through structured slots and values, including corrections to earlier values \citep{jacqmin2022you, mcg}. The Frames corpus models multiple goal frames and references among competing alternatives, while conversational recommender systems support iterative preference refinement through user critiques \citep{el2017frames,schulz2017frame,mcg,jannach2021survey}. These settings are closely related, but primarily evaluate state recovery, response generation, or recommendation efficiency. IntentDelta instead grounds each change in task results and separately evaluates both the maintained intention state and the downstream behavior of a task-executing agent.

\section{IntentDelta Benchmark}
IntentDelta evaluates whether agents can maintain and act on user intentions as constraints and priorities evolve. We formulate the agent's and user's decision processes, and model evolving intentions and trade-offs. We then define metrics for intention recovery and action execution.

\subsection{Task Formulation}
We describe this idealized interaction through two coupled decision processes: an agent process and a user process. This conceptual description motivates the benchmark design.

\paragraph{Agent process.}
We characterize the agent-side interaction as a partially observable decision process 
\((\mathcal{S}^A, \mathcal{A}^A, \mathcal{O}^A, T^A)\), 
where \(\mathcal{S}^A\) is the state space, \(\mathcal{A}^A\) is the agent's task action space, \(\mathcal{O}^A\) is the observation space, and \(T^A\) is the transition function. The observation space consists of information from both the task environment and the user, such that 
\(\mathcal{O}^A = \mathcal{O}^A_{\mathrm{env}} \cup \mathcal{O}^A_{\mathrm{user}}\).
For example, \(\mathcal{O}^A_{\mathrm{env}}\) may contain product search results or available travel options, while \(\mathcal{O}^A_{\mathrm{user}}\) contains the user's utterances. The user's current intention \(I_t\) is part of the underlying state but is not directly observable to the agent.

\paragraph{User process.}
We characterize the user-side interaction as a Markov decision process 
\((\mathcal{S}^U, \mathcal{A}^U, T^U)\).
A user state \(s_t^U = (I_t, C_t)\) consists of the current intention \(I_t\) and the interaction context \(C_t\). For example, in a travel-planning task, \(I_t\) may specify a \$1,800 budget and a preference for an earlier return, while \(C_t\) contains the relevant flight and itinerary information together with the agent's latest result. The user's action is an utterance \(u_t \in \mathcal{A}^U\); after seeing that an earlier flight conflicts with the original plan, the user may say to the agent, ``I can stretch the budget a bit if that gets me home earlier.''

This idealized description captures how user intentions
may evolve in response to task outcomes. During dataset
construction, we ground intention updates in human
exploration of the task environment. During evaluation,
we replay the resulting fixed user utterance sequence
across agents. Subsequent utterances do not depend on
the tested agent's actions, while each agent independently
interacts with the environment. This protocol enables
controlled comparisons under the same evolving requirements.

\subsection{Modeling Evolving User Intention}

\textbf{Representing intention as constraints.} 
We represent user intention as a set of task-relevant constraints that make an abstract goal concrete and assessable, such as a budget below \$500, a required hotel rating, or a preferred return time. This abstraction is grounded in prior dialogue-state representations and empirical studies showing that users progressively articulate and revise their requirements through interaction \cite{zhu-etal-2020-crosswoz, Cai2019TowardsAT, vaithilingam2025semantic, wang2026trajectory2task}. At turn \(t\), the intention \(I_t\) consists of the currently active constraints together with their values, scopes, and priorities, allowing us to characterize how user requirements and trade-offs evolve over time.

\textbf{Conditions for intention change.}
During dataset construction, intention changes are grounded in the interaction context \(C_t\), including environment observations and human reference outcomes. These results provide information that can motivate a revision to the current intention. We consider two common conditions for such changes. First, \textbf{real-world feasibility} may reveal that previously stated requirements cannot be jointly satisfied. For example, a user searching for a laptop may initially request high-end performance while maintaining a budget that is incompatible with the available products, leading the user to relax the budget or compromise on another requirement. Second, \textbf{user preference revision} may occur after the user observes concrete alternatives. A user who initially prioritizes performance, for instance, may decide after comparing several laptops that portability or battery life is more important for their actual use case. In both cases, reference exploration provides information that can motivate a transition from \(I_t\) to \(I_{t+1}\).

\textbf{Priority and trade-offs.}
A defining feature of IntentDelta is that user decision making often requires resolving trade-offs among competing constraints. Such trade-offs are central to realistic decision making but are often underrepresented in existing agent benchmarks, where constraints are typically treated as equally important or independently satisfiable. To capture these trade-offs, we associate each constraint with one of three priority levels: \textbf{must-have}, \textbf{preferred}, and \textbf{optional}. Must-have constraints represent requirements that should be preserved whenever feasible, preferred constraints are desirable but may be sacrificed when necessary, and optional constraints represent lower-priority preferences. When no available solution satisfies all active constraints, these priorities specify which requirements the user is more willing to compromise on. Importantly, priority is itself part of the evolving intention state and may change as the user observes new outcomes. More generally, an intention update may introduce or remove a requirement, revise its value, or change its priority.

\begin{comment}
\textbf{User utterances.}
The agent does not directly observe the intention state \(I_t\), but only the user's utterance \(u_t\). An utterance may express an intention update explicitly, such as stating that hotel quality matters more than a convenient flight time, or implicitly, with the intended change inferred from the preceding interaction context. For example, after first planning to drive under a \$1,500 budget and later considering a more expensive flight-based plan, a user may simply say, ``Actually, let's drive after all,'' without restating the earlier budget constraint. IntentDelta therefore varies how intention changes are communicated while keeping the underlying intention state well defined.
\end{comment}

\subsection{Evaluation Metrics}
\label{sec:metrics}
Our task formulation distinguishes the user's latent intention $I_t$ from the agent's resulting actions. Accordingly, we evaluate agents at two complementary levels: \textit{intention recovery} and \textit{action execution}. The former measures agents' intention recognition, while the latter evaluates agents' task execution.

\paragraph{Intention-Level Evaluation.}
At each turn, we semantically match the agent's predicted constraints with the gold active constraints, then evaluate their content and priority separately. We report four metrics: \textit{Detection} ($\frac{\#\text{detected gold constraints}}{\#\text{gold constraints}}$) measures the proportion of gold constraints identified by the agent. \textit{Content Accuracy} ($\frac{\#\text{correct contents}}{\#\text{detected constraints}}$) measures whether detected constraints have correct values and scopes. \textit{Priority Accuracy} ($\frac{\#\text{correct priorities}}{\#\text{detected constraints}}$) measures whether agents correctly distinguish must-have, preferred, and optional requirements. Finally, \textit{Intention Recovery} ($\frac{\#\text{fully recovered constraints}}{\#\text{gold constraints}}$) measures whether constraints are recovered with both correct content and priority, serving as our primary intention-level metric. We additionally report prediction precision and obsolete constraint
retention in Appendix~\ref{app:precision_retention} to assess extra
predictions and failures to discard superseded requirements.

\paragraph{Action-Level Evaluation.}
We evaluate whether the agent's actions satisfy the gold intention under environmental constraints. Following our priority model, must-have constraints define the minimum conditions for an acceptable outcome, while preferred and optional constraints distinguish the quality of acceptable outcomes. We therefore report two metrics: \textit{Hard Success} and \textit{Action Score}.

\textit{Hard Success} evaluates satisfaction of must-have constraints. Human annotators execute each task to produce a gold trajectory and assess task feasibility. A task is considered \textit{feasible} if all must-have constraints can be jointly satisfied, and \textit{infeasible} otherwise. For feasible tasks, the agent must satisfy all must-have constraints; for infeasible tasks, it must satisfy at least as many as the human execution. Let $M_a$ and $M_g$ be the numbers of must-have constraints satisfied by the agent and gold trajectory, respectively, and $n_M$ the total number of must-have constraints:

\begin{equation}
G_t =
\begin{cases}
\mathbf{1}[M_a = n_M], & \text{if feasible},\\
\mathbf{1}[M_a \geq M_g], & \text{if infeasible}.
\end{cases}
\end{equation}

\textit{Action Score} provides a more fine-grained evaluation by combining must-have satisfaction with lower-priority preferences. Since satisfying soft constraints cannot compensate for failing must-have requirements, the score is conditioned on Hard Success. We assign weights of 2 and 1 to preferred and optional constraints, respectively, yielding $S_a = 2P_a + O_a$ and $S_g = 2P_g + O_g$ for the agent and gold trajectories. At each turn, we compute $\mathrm{ActionScore}_t = G_t \cdot \min(1, S_a/S_g)$, which rewards soft-constraint satisfaction relative to human execution only when Hard Success is achieved. When $S_g = 0$, we set $\mathrm{ActionScore}_t = G_t$. The 2:1 weighting distinguishes preferred from optional constraints without assuming a precise user utility function. Appendix~\ref{app:weight_sensitivity} examines the sensitivity of model rankings to alternative weightings. Both metrics are computed at each turn and averaged across turns for reporting.

\section{Benchmark Construction}

\subsection{Domain}
We build IntentDelta upon two existing benchmarks, WebShop \cite{yao2022webshop} and TravelPlanner \cite{xie2024travelplanner}, reusing their task environments and original user requests as initial tasks. WebShop provides crowdsourced shopping instructions grounded in real-world product data, where agents search, inspect, and select products based on attributes, prices, and specification options. TravelPlanner provides synthesized travel requests with human-annotated reference plans, supported by structured information and search tools for transportation, accommodations, restaurants, and attractions. These two domains offer complementary settings for evolving intentions: WebShop focuses on product-level requirements, while TravelPlanner involves interdependent constraints across multiple itinerary components, where changes to one requirement may affect the feasibility of others.

\subsection{Human Construction}
\textbf{Human Trajectory Creation. }Members of the author team manually construct 638 trajectories based on the initial tasks from WebShop and TravelPlanner, with each trajectory taking approximately 5–10 minutes. At each turn, authors construct three components: (1) a structured intention containing constraints and priorities, where authors consider potential trade-offs based on previous task outcomes and determine which requirements are must-have, preferred, or optional; (2) a natural-language user utterance that communicates the updated intention to the agent; and (3) a human reference execution that reflects the current intention. Authors have full access to the task environment, allowing them to independently explore available options and construct reference executions. Further details of the annotation procedure are provided in
Appendix~\ref{app:annotation}, with the WebShop and TravelPlanner
interfaces shown in Figures~\ref{fig:webshop_annotation}
and~\ref{fig:travelplanner_annotation}, respectively.

\textbf{Two-Stage Annotation Validation and Quality Assurance.} Since user intention reflects a subjective mental model, the same utterance or trade-off may admit multiple interpretations. To minimize ambiguity and ensure consistent interpretation, we conduct a two-stage cross-verification process. First, each trajectory is cross-verified by an author who did not construct it. Reviewers examine the consistency of the user utterances, cumulative intentions, priorities, and reference actions. Ambiguous or unreasonable cases are discussed and revised until consensus is reached. Second, we sample 20\% of the dataset (128 trajectories) for independent annotation by external annotators. External reviewers assess the consistency and clarity of the three annotated components (intentions, utterances, and executions) to identify potential inconsistencies or ambiguities, achieving 97\% agreement with the authors' annotations.

\subsection{Dataset Characteristics}
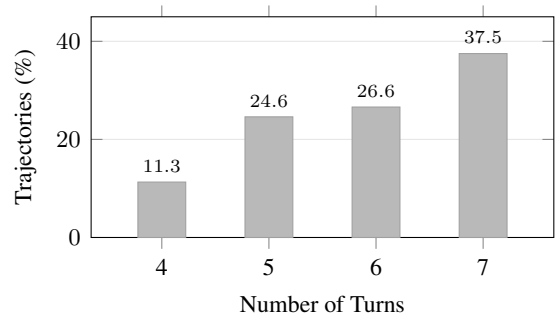
\begin{figure}[t]
\centering
\begin{tikzpicture}
\begin{axis}[
    ybar,
    width=\columnwidth,
    height=4.5cm,
    bar width=18pt,
    ymin=0,
    ymax=45,
    ylabel={Trajectories (\%)},
    xlabel={Number of Turns},
    symbolic x coords={4,5,6,7},
    xtick=data,
    enlarge x limits=0.22,
    ymajorgrids=true,
    grid style={gray!20},
    nodes near coords,
    every node near coord/.append style={
        font=\scriptsize
    },
    tick label style={font=\small},
    label style={font=\small},
]
\addplot[
    fill=gray!55,
    draw=gray!75
] coordinates {
    (4,11.3)
    (5,24.6)
    (6,26.6)
    (7,37.5)
};
\end{axis}
\end{tikzpicture}
\caption{Distribution of trajectory lengths across
WebShop and TravelPlanner, showing the percentage
of trajectories with different numbers of turns.}
\label{fig:trajectory_length}
\end{figure}

IntentDelta contains 638 trajectories and 3,766 turns across WebShop and TravelPlanner. Figure~\ref{fig:trajectory_length} presents the distribution of trajectory lengths, while Table~\ref{tab:dataset_characteristics} summarizes intention change characteristics. Reprioritization accounts for approximately 46\% of change events in both domains. Compound changes occur in 66.5\% of changed turns in WebShop and 81.5\% in TravelPlanner, highlighting the prevalence of simultaneous constraint revisions. Feasibility-driven changes account for 45.1\% and 26.8\% of changed turns, respectively.

\begin{table}[t]
\centering
\small
\begin{tabular}{lcc}
\toprule
\textbf{Characteristics} &
\textbf{WebShop} &
\textbf{TravelPlanner} \\
\midrule
\multicolumn{3}{l}{\textit{Change Operations (\% of events)}} \\
Add                 & 20.4 & 37.3 \\
Relax               & 18.0 & 8.3  \\
Override            & 15.0 & 8.5  \\
Reprioritize        & 46.6 & 46.0 \\
\midrule
\multicolumn{3}{l}{\textit{Change Triggers (\% of change turns)}} \\
Feasibility-driven  & 45.1 & 26.8 \\
Preference-driven   & 54.9 & 73.2 \\
\midrule
\multicolumn{3}{l}{\textit{Change Complexity (\% of change turns)}} \\
Compound changes    & 66.5 & 81.5 \\
\bottomrule
\end{tabular}
\caption{Intention change characteristics across
WebShop and TravelPlanner. Operation percentages
are calculated over change events, while trigger
and compound-change percentages are calculated
over turns containing intention changes.}
\label{tab:dataset_characteristics}
\end{table}

\section{Experiments}
We evaluate LLM agents on evolving user intentions in WebShop and TravelPlanner, reporting overall performance and analyzing intention tracking and execution failures.
\begin{table*}[!t]
\centering
\small
\begin{threeparttable}
\setlength{\tabcolsep}{4.2pt}
\renewcommand{\arraystretch}{1.13}

\begin{tabular*}{\textwidth}{@{\extracolsep{\fill}}lrrrrrrrr@{}}
\toprule
& \multicolumn{4}{c}{\textbf{TravelPlanner}}
& \multicolumn{4}{c}{\textbf{WebShop}} \\
\cmidrule(lr){2-5}
\cmidrule(l){6-9}
& \multicolumn{2}{c}{Intention} & \multicolumn{2}{c}{Action}
& \multicolumn{2}{c}{Intention} & \multicolumn{2}{c}{Action} \\
\cmidrule(lr){2-3}\cmidrule(lr){4-5}\cmidrule(lr){6-7}\cmidrule(l){8-9}
\textbf{Model}
& Det.\,$\uparrow$ & Rec.\,$\uparrow$ & \shortstack{Hard\\Succ.\,$\uparrow$} & \shortstack{Action\\Score\,$\uparrow$}
& Det.\,$\uparrow$ & Rec.\,$\uparrow$ & \shortstack{Hard\\Succ.\,$\uparrow$} & \shortstack{Action\\Score\,$\uparrow$} \\
\midrule
\texttt{gpt-5.6-sol}      & 70.4 & 38.4 & 52.3 & 49.6 & 92.2 & 50.6 & \textbf{58.6} & \textbf{49.0} \\
\texttt{gpt-5.6-luna}     & \textbf{72.1} & 38.2 & 46.5 & 42.3 & 91.6 & 49.4 & 54.6 & 45.1 \\
\midrule
\texttt{claude-opus-4.8}  & 68.4 & 37.3 & 47.0 & 42.3 & 89.2 & \textbf{52.0} & 52.0 & 41.9 \\
\texttt{claude-sonnet-5}  & 70.8 & \textbf{39.1} & 55.6 & 51.6 & 90.6 & 50.6 & 52.0 & 42.7 \\
\midrule
\texttt{deepseek-r1}      & 43.1 & 23.4 & 36.2 & 28.9 & 66.0 & 35.6 & 48.0 & 38.6 \\
\addlinespace[1.7pt]
\texttt{grok-4.6}         & 68.4 & 37.3 & \textbf{60.3} & \textbf{57.1} & \textbf{92.8} & 51.4 & 58.2 & 47.2 \\
\addlinespace[1.7pt]
\texttt{kimi-k2.5}        & 69.6 & 36.2 & 31.9 & 26.6 & 89.0 & 49.3 & 52.1 & 42.9 \\
\addlinespace[1.7pt]
\texttt{minimax-m2.5}     & 58.2 & 31.2 & 25.4 & 21.6 & 75.1 & 40.1 & 49.9 & 40.5 \\
\addlinespace[1.7pt]
\texttt{nova-pro}         & 49.4 & 25.6 & 10.0 &  6.6 & 65.9 & 32.5 & 41.9 & 34.0 \\
\midrule
\texttt{qwen3-235b}       & 57.9 & 30.4 & 17.2 & 13.3 & 80.8 & 40.4 & 47.6 & 38.8 \\
\texttt{qwen3-32b}        & 53.0 & 27.5 & 11.9 &  7.7 & 80.5 & 40.1 & 37.8 & 30.7 \\
\midrule
\textit{Average}          & 61.9 & 33.1 & 35.8 & 31.6 & 83.1 & 44.7 & 50.2 & 41.0 \\
\bottomrule
\end{tabular*}

\caption{
Main results on TravelPlanner and WebShop. Det. and Rec. denote Detection and Intention Recovery, respectively. All values are on a 0--100 scale; higher is better. \textbf{Bold} indicates the highest numerical value in each column.
}
\label{tab:main-results}

\end{threeparttable}
\end{table*}

\subsection{Setup}

\noindent\textbf{Models.} We evaluate eleven models: GPT-5.6 Sol and Luna; Claude Opus 4.8 and Sonnet 5; DeepSeek R1; Grok 4.6; Kimi K2.5; MiniMax M2.5; Nova Pro; and Qwen3-235B and Qwen3-32B.

\noindent\textbf{Agent and tools.} All models use a common ReAct framework \cite{yao2023react}, interleaving reasoning, actions, and environment observations. We reuse the search tools and data resources provided by WebShop and TravelPlanner. Alongside its actions, each agent predicts the complete active intention as constraints with fields, values, and priorities. Execution details and prompts are provided in Appendix~\ref{app:execution}.

\noindent\textbf{Inference settings.}
All agents use each model's default decoding configuration. We do not explicitly set temperature, sampling parameters, or reasoning budgets; reasoning models therefore use their default reasoning effort. TravelPlanner agents may take up to 8 steps per turn, with up to 12 search calls per step and a 12k-token output limit, increased to 48k if the response is truncated. WebShop agents may take up to 20 environment actions per turn, with an approximately 8k-token output limit.

\noindent\textbf{Evaluation.} We evaluate all models on 304 TravelPlanner trajectories (1,946 turns) and 334 WebShop trajectories (1,820 turns), replaying the same annotated user utterance sequence for every model. These updates are grounded in annotators' independent exploration and reference executions in the shared environment, enabling controlled comparisons under the same evolving requirements. Agents receive the user history and public environment observations; gold annotations and human reference outcomes are withheld. All LLM judgments are produced by GPT-5.6 Sol, with one judgment per item, default decoding, and a 16k-token output limit. GPT-5.6 Sol is also an evaluated model. The judge assesses constraint matching and satisfaction against fixed, human-derived criteria and environment evidence, using human reference executions for outcome comparison. Model identities are withheld from judge prompts, and all metrics are computed deterministically in code following Section~3.3. We validate LLM judgments against human evaluations on 100 sampled agent outputs, achieving 95\% agreement. Evaluation details and judging prompts are provided in Appendix~C.

\subsection{Main Results}
Table~\ref{tab:main-results} presents the overall performance of 11 evaluated models across TravelPlanner and WebShop. Models achieve an average intention recovery of 33.1\% in TravelPlanner and 44.7\% in WebShop, while average Hard Success reaches only 35.8\% and 50.2\%, and average Action Score 31.6\% and 41.0\%, respectively. These results reveal substantial room for improvement in both maintaining evolving user intentions and executing actions that satisfy them.

Model performance also differs between intention recovery and action execution. In TravelPlanner, Claude Sonnet 5 achieves the highest intention recovery (39.1\%), whereas Grok 4.6 obtains the highest Hard Success (60.3\%) and Action Score (57.1\%). Similarly, in WebShop, Claude Opus 4.8 performs best in intention recovery (52.0\%), while GPT-5.6 Sol achieves the highest Hard Success (58.6\%) and Action Score (49.0\%). No single model consistently leads both evaluation levels, motivating a more detailed analysis of intention tracking and execution failures. Model rankings are robust to the tested alternative soft-constraint weightings: Kendall's $\tau$ is at least 0.964 in both domains, and no model changes rank by more than one position (Appendix~\ref{app:weight_sensitivity}).

\subsection{Research Challenge Analysis}
\begin{figure}[t]
\centering
\begin{tikzpicture}
\begin{axis}[
    width=\columnwidth, height=4.3cm,
    ymin=0, ymax=80,
    ylabel={Intention recovery (\%)},
    xlabel={Turns since the constraint was introduced},
    symbolic x coords={1,2--3,4+},
    xtick=data,
    enlarge x limits=0.25,
    ymajorgrids=true, grid style={gray!20},
    legend style={font=\scriptsize, at={(0.98,0.98)}, anchor=north east},
    tick label style={font=\small}, label style={font=\small},
]
\addplot[mark=*, thick, blue!70, nodes near coords, every node near coord/.append style={font=\scriptsize, anchor=north, yshift=-2pt}] coordinates {(1,58.7) (2--3,31.6) (4+,27.6)};
\addlegendentry{TravelPlanner}
\addplot[mark=square*, thick, orange!80!black, nodes near coords, every node near coord/.append style={font=\scriptsize, anchor=south, yshift=2pt}] coordinates {(1,63.1) (2--3,34.8) (4+,32.7)};
\addlegendentry{WebShop}
\end{axis}
\end{tikzpicture}
\caption{Intention recovery of carried-over constraints (not revised in the current turn) by how many turns ago they were introduced. Macro-average over 11 models; the decline holds for all 11 models in both domains.}
\label{fig:constraint_age}
\end{figure}
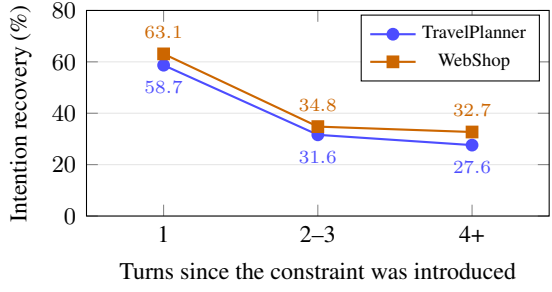

\begin{figure}
\centering
\includegraphics[width=0.5\textwidth]{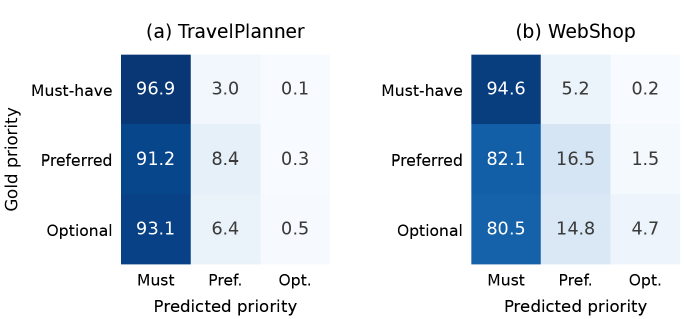}
\caption{Priority prediction in (a) TravelPlanner and (b) WebShop. Each row shows the distribution of predicted priorities for a gold priority, normalized to percentages. Diagonal entries indicate correct predictions.}
\label{fig:priority_downgrade}
\end{figure}

\begin{figure}
    \centering
    \includegraphics[width=\linewidth]{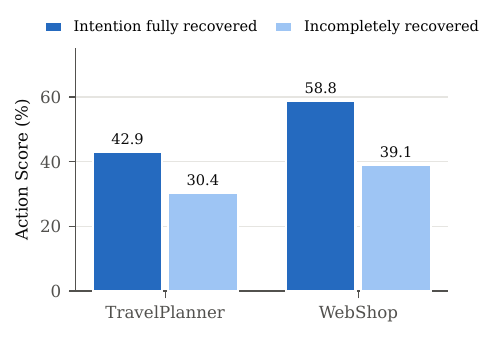}
    \caption{Action Score for turns with fully versus incompletely recovered user intentions in TravelPlanner and WebShop. Full intention recovery is associated with higher Action Scores in both domains.}
    \label{fig:recovery_action_score}
\end{figure}

\paragraph{Agents recover only a minority of active constraints.}
Average Detection reaches 61.9\% in TravelPlanner and 83.1\% in WebShop, with Content Accuracy of 88.1\% and 83.7\%, respectively. However, Priority Accuracy is only 58.9\% and 62.8\%, and Intention Recovery, which requires both correct content and priority, is 33.1\% and 44.7\%. Recovery is limited by three factors: missed constraints, incorrect content, and, most prominently, incorrect priority. When multiple constraints are revised within the same turn, intention recovery drops from 52.5\% to 28.7\% in TravelPlanner and from 52.4\% to 32.3\% in WebShop compared with single-constraint changes. The degradation is consistent across all 11 models in both domains.

Earlier constraints are harder to recover. Among constraints that are carried over unchanged from earlier turns, intention recovery decreases with constraint age, as shown in Figure~\ref{fig:constraint_age}. In TravelPlanner, recovery drops from 58.7\% for constraints introduced one turn earlier to 27.6\% for constraints introduced at least four turns earlier; in WebShop, it drops from 63.1\% to 32.7\%. This suggests that evolving-intent interaction requires not only incorporating new revisions, but also maintaining requirements that remain active across turns.

\paragraph{Agents treat soft preferences as hard requirements.}
Agents struggle to distinguish constraint priorities, achieving only 58.9\% Priority Accuracy in TravelPlanner and 62.8\% in WebShop. As shown in Figure~\ref{fig:priority_downgrade}(a--b), agents incorrectly classify 91.2\% and 82.1\% of preferred constraints, and 93.1\% and 80.5\% of optional constraints, as must-have in TravelPlanner and WebShop, respectively. Such misunderstandings of priority can lead agents to make incorrect trade-offs during execution.

\paragraph{Intention recovery is associated with, but does not guarantee, successful action.} Better intention recovery is associated with better execution (Figure~\ref{fig:recovery_action_score}). Action Score reaches 42.9\% in TravelPlanner and 58.8\% in WebShop when the intention is fully recovered, compared with 30.4\% and 39.1\% when it is incompletely recovered. However, correct intention recovery does not guarantee a correct action. In TravelPlanner, among failed turns in which the intention is fully recovered, 79.4\% violate the budget, suggesting difficulty accounting for costs across multiple itinerary components, and 33.1\% book accommodation that violates its minimum-stay or capacity rules, a requirement the user never states. In WebShop, 60.4\% of such failures violate a product attribute and 41.7\% a product option such as size, color, or quantity, indicating failures in retrieving the appropriate product or selecting the correct option despite correctly representing the user's requirements. A breakdown of constraint violations and invalid bookings in failed turns is provided in Appendix~\ref{app:error_analysis}, Table~\ref{tab:failure_by_intention_recovery}.

\section{Discussion}

Successful task execution requires both valid actions and an accurate understanding of what the user intends to achieve \cite{bansal2024challenges}. Our benchmark examines this connection by evaluating whether agents maintain and act on user intentions as constraints and priorities evolve. The results expose difficulties in (1) revising intentions while preserving earlier requirements, (2) understanding trade-offs, and (3) executing the resulting constraints. These challenges motivate studying how agents interpret revisions within their context. This connects to prior work on preference trade-offs \cite{10.1145/1639714.1639754} and memory updates \cite{wu2024longmemeval}. Future approaches could jointly consider user feedback, dialogue history, and environment state to determine which requirements have changed and which remain active.

When context leaves the user’s intention uncertain, agents could ask targeted clarification questions. Prior work aims to maximize information gain to determine when and what to ask for clarification \cite{wang2025learning, kobalczyk2025active, zhou2026should}. Our benchmark further specifies what this information consists of, including constraint content, scope, priority, and changes across turns, providing finer-grained targets for deciding what to ask as user intentions evolve. 

\section{Conclusion}

We introduced IntentDelta, a multi-turn benchmark for evaluating agents under evolving user constraints and priorities in WebShop and TravelPlanner. Human-constructed trajectories pair natural-language updates with cumulative intention annotations and reference actions, enabling separate evaluation of intention recovery and action quality. Across eleven models, the results reveal incomplete recovery of active requirements, frequent priority errors, and execution failures despite higher intention recovery. These findings motivate improved methods for maintaining current requirements, updating their priorities, and verifying actions against environment evidence.

\section{Limitations}
IntentDelta uses human-authored trajectories in two task environments, which may not fully represent natural user interactions or generalize to other domains. For infeasible tasks, Hard Success uses the human reference as a comparison threshold ($M_a \geq M_g$). The reference captures one plausible compromise among competing requirements, while other reasonable solutions may satisfy different subsets of constraints. Consequently, this metric depends on the selected reference and does not fully capture the quality of alternative trade-offs. Three discrete priority tiers also simplify preferences that may be contextual or continuous. 

\bibliography{custom}

\appendix

\section{Human Annotation}
\label{app:annotation}

\begin{figure*}[p]
    \centering
    \includegraphics[
        width=\textwidth,
        height=0.40\textheight,
        keepaspectratio
    ]{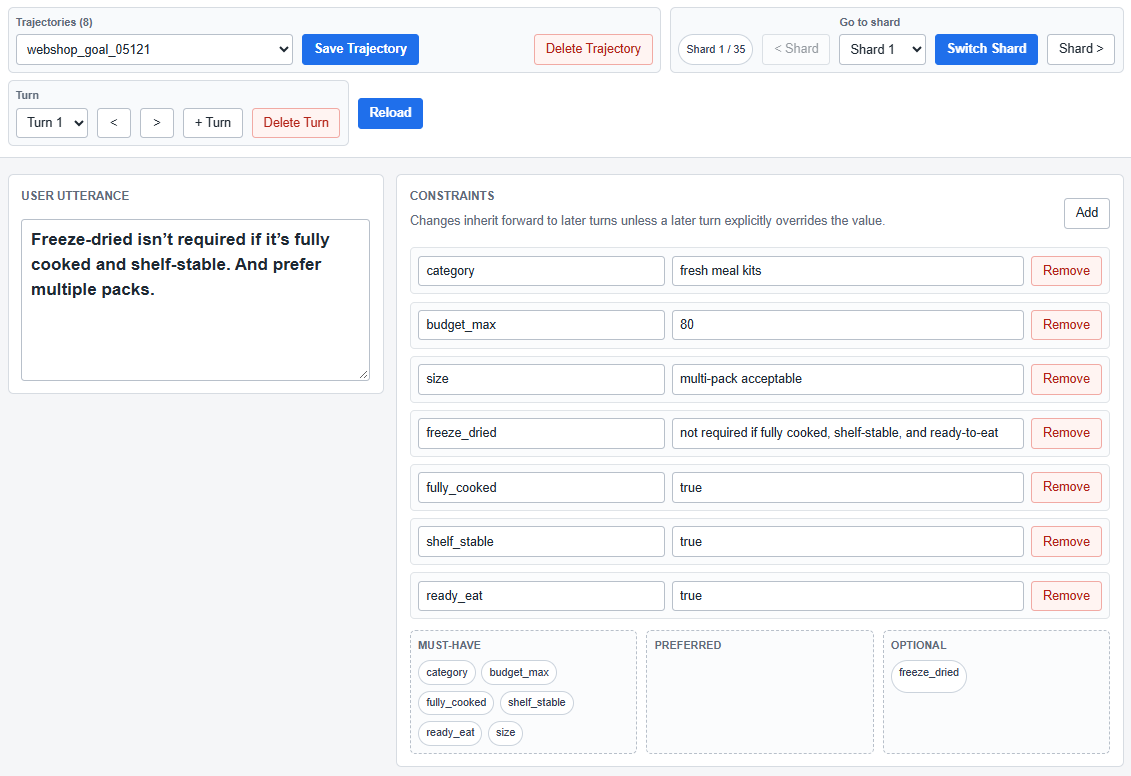}

    \smallskip
    {\small (a) User utterance, cumulative constraints,
    and priority assignments.\par}

    \medskip
    \includegraphics[
        width=\textwidth,
        height=0.40\textheight,
        keepaspectratio
    ]{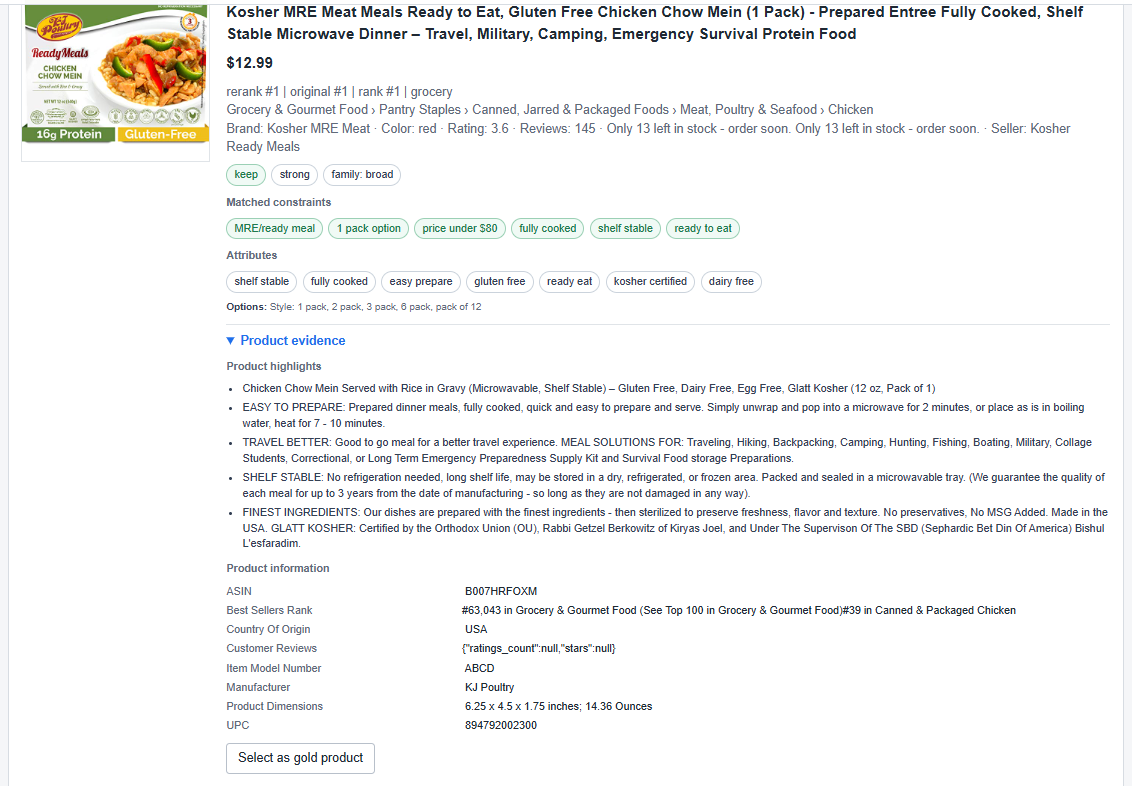}

    \smallskip
    {\small (b) Candidate product and supporting evidence.\par}

    \caption{
    WebShop annotation interface, illustrated with a ready-to-eat
    meal task.
    (a) Annotators edit user utterances, cumulative constraints,
    and their priorities.
    (b) Annotators inspect product information, available options,
    and supporting evidence before selecting a reference product.
    }
    \label{fig:webshop_annotation}
\end{figure*}

\begin{figure*}[p]
    \centering
    \includegraphics[
        width=\textwidth,
        height=0.40\textheight,
        keepaspectratio
    ]{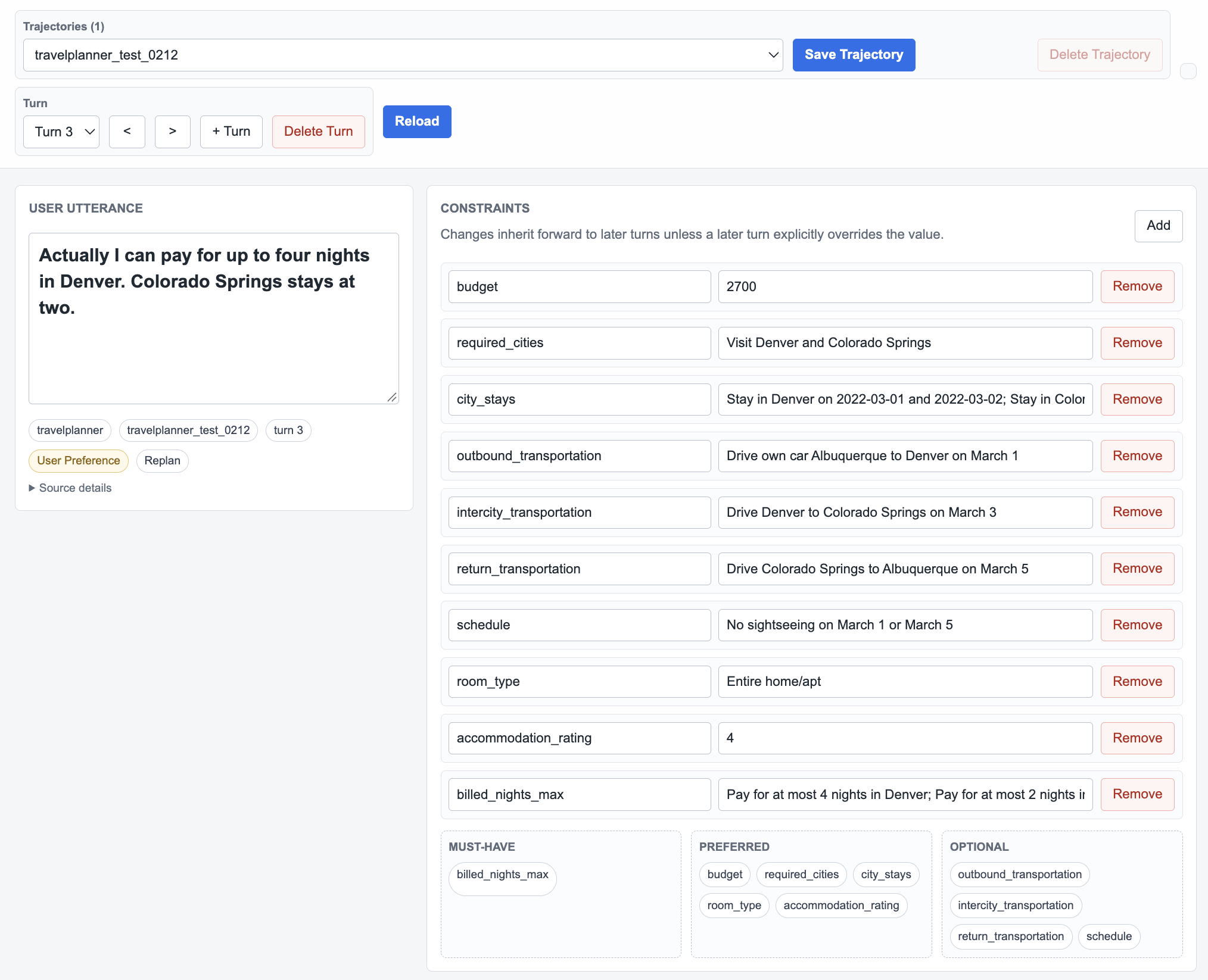}

    \smallskip
    {\small (a) User utterance, cumulative constraints,
    and priority assignments.\par}

    \medskip
    \includegraphics[
        width=\textwidth,
        height=0.40\textheight,
        keepaspectratio
    ]{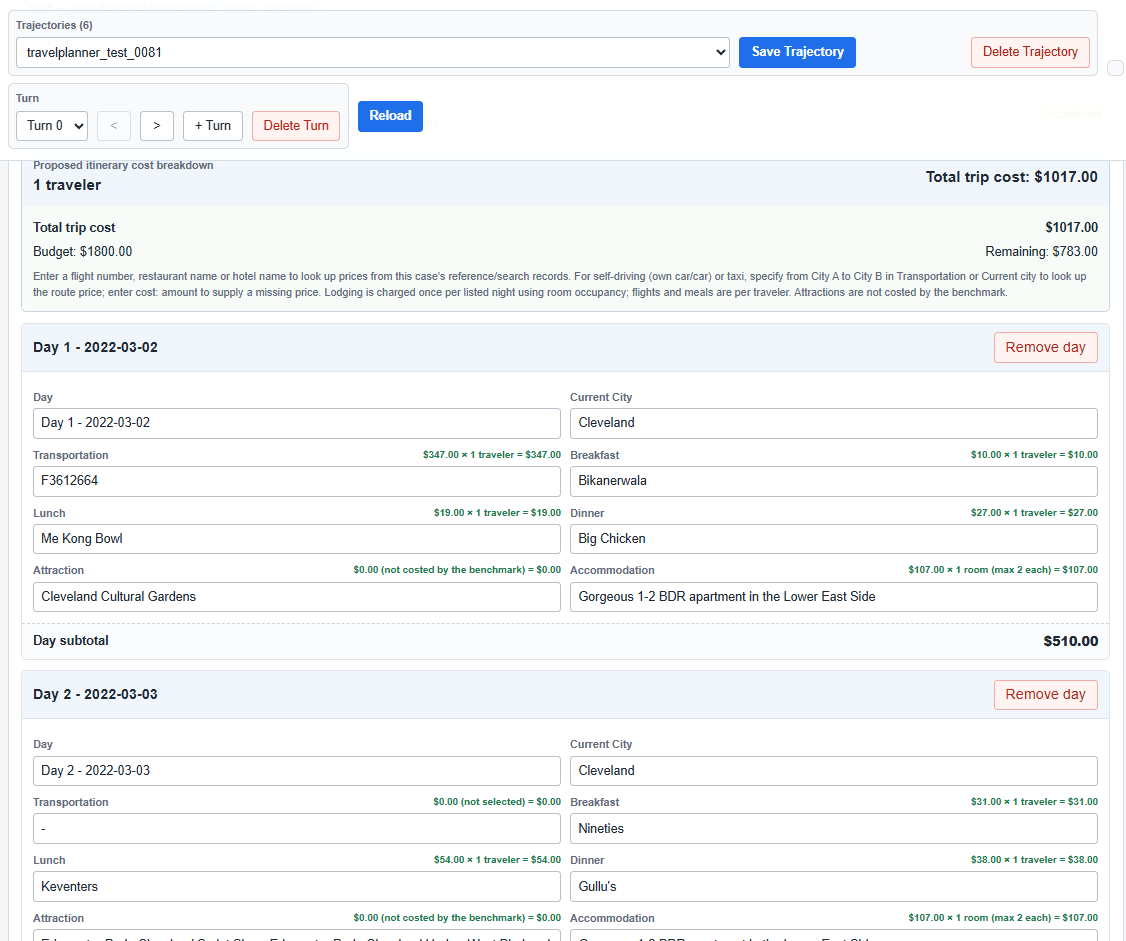}

    \smallskip
    {\small (b) Reference itinerary editor and cost breakdown.\par}

    \caption{
    TravelPlanner annotation interface.
    (a) Annotators record a location-specific update allowing
    up to four billed accommodation nights in Denver while
    retaining a two-night limit in Colorado Springs, alongside
    other active constraints and their priorities.
    (b) Annotators edit a daily reference itinerary and inspect
    itemized costs and the total trip cost, illustrated with
    a Cleveland itinerary.
    The two panels show different trajectories.
    }
    \label{fig:travelplanner_annotation}
\end{figure*}

Our annotation interface supports the construction and review of
multi-turn trajectories in WebShop and TravelPlanner. Each turn
contains a natural-language user utterance, the current cumulative
intention state, and a human reference action. Annotators inspect
environment evidence alongside these components to check consistency
between what the user communicates, what the user intends, and what
the reference execution achieves.
Figures~\ref{fig:webshop_annotation}(a)
and~\ref{fig:travelplanner_annotation}(a) show the utterance,
constraint, and priority editors.
Figure~\ref{fig:webshop_annotation}(b) shows product evidence
used to review a reference selection, while
Figure~\ref{fig:travelplanner_annotation}(b) shows the itinerary
editor and cost breakdown. 

\paragraph{Ethics approval.}
The data collection and annotation protocol was approved by our
institutional review board.

\paragraph{Recruitment.}
For the external validation stage described in Section~4.2,
we recruited university students through word of mouth and
recruitment posters. These participants validated sampled
trajectories constructed by the author team. Participants provided
informed consent through a consent form before beginning annotation
and received \$15 for every 30 cases annotated.

\paragraph{Utterances and cumulative intentions.}
Annotators can edit the user utterance and add, revise, or remove
structured constraints. The intention state records all currently
active requirements, including those carried over from previous turns.
Constraints are assigned to three priority tiers:
\textit{Must-have}, \textit{Preferred}, and \textit{Optional}.
The interface displays these tiers as separate columns and supports
drag-and-drop reassignment.
Constraint edits propagate to subsequent turns where applicable,
while preserving explicit later overrides. Annotators can navigate
between turns to review the resulting trajectory and save all changes
at the trajectory level.

\paragraph{Environment evidence and reference actions.}
For WebShop, annotators inspect candidate products, prices,
descriptions, attributes, and available options. A reference action
specifies a selected product and its exact option values.

For TravelPlanner, annotators inspect transportation, accommodation,
restaurant, and attraction records and edit a day-by-day reference
itinerary. Each day specifies its location, transportation, meals,
attractions, and accommodation. The interface also provides a cost
breakdown based on the available records, including party size,
accommodation occupancy, and minimum-night billing.
The displayed breakdown describes billing assumptions;
minimum-stay validity is assessed separately using the nights
explicitly represented in the itinerary.
Components not costed by the benchmark are explicitly marked.

\paragraph{Feasibility and trade-offs.}
Annotators record whether the active must-have requirements
are jointly feasible in the reviewed environment and may
document unavoidable trade-offs. Reference actions represent
concrete human choices under these conditions and need not
satisfy every active requirement.

\paragraph{Illustrative examples.}
Figure~\ref{fig:webshop_annotation}(a) shows utterance editing,
active constraints, and priority assignments for a ready-to-eat
meal task. Figure~\ref{fig:webshop_annotation}(b) shows candidate
product evidence used to review a reference selection.
Figure~\ref{fig:travelplanner_annotation}(a) illustrates a
location-specific relaxation: the user allows up to four billed
accommodation nights in Denver while retaining a two-night limit
in Colorado Springs. This example highlights the need to preserve
the scope of an update rather than applying it to the entire itinerary.
Figure~\ref{fig:travelplanner_annotation}(b) shows reference itinerary
editing and cost inspection for a separate Cleveland trip.

\section{Agent Execution and Prompts}
\label{app:execution}

\subsection{Execution Protocol}
\paragraph{Fixed-user interaction protocol.}
During dataset construction, annotators formulate intention updates in response to outcomes obtained along a human reference trajectory. During evaluation, we replay the resulting user utterances in their original order, using the same utterance sequence for every evaluated agent. At each turn, the agent infers the cumulative intention from the available user history and interacts with the task environment. The gold intention, human priorities, feasibility annotation, and reference action are withheld from the execution agent. Result-grounded intention evolution describes dataset construction. Evaluation replays a shared sequence of annotated user updates rather than generating new user responses to each tested agent's actions.

\paragraph{Intention prediction.}
The agent reports the complete active intention, including
requirements carried over from earlier turns, using the schema
in Appendix~\ref{app:shared_intent_prompt}. Each requirement includes its value, priority, and any necessary traveler, location, or time context.

\paragraph{Domain-specific execution.}
In WebShop, the agent observes the current page and can search,
inspect products, select options, navigate between result pages,
and finalize a product selection. Its prompt includes the user
utterance history and a concise trace of earlier internal steps
within the current turn.

In TravelPlanner, the agent can invoke \texttt{CitySearch},
\texttt{FlightSearch}, \texttt{AccommodationSearch},
\texttt{RestaurantSearch}, \texttt{AttractionSearch}, and
\texttt{GoogleDistanceMatrix}. Retrieved information is recorded through \texttt{NotebookWrite}, and the agent submits a complete itinerary through \texttt{Planner}. The underlying reference database is exposed through tool observations rather than directly supplied as gold information. Execution proceeds within a configured internal-step budget and records the intention prediction, actions, observations, and termination reason. The WebShop and TravelPlanner codebases are licensed under MIT, and the TravelPlanner dataset is licensed under CC BY 4.0.

The execution prompts combine shared intention-representation
instructions and an action-selection policy with domain-specific
interaction rules. We present the shared blocks once, followed
by the two domain-specific templates. Static instruction wording
is preserved, with line wrapping adjusted for presentation.
Named placeholders indicate where shared blocks or serialized
runtime inputs are inserted; they are not sent literally to
the model.

\subsection{Shared Intention Representation}
\label{app:shared_intent_prompt}

Both domains use the following intention schema. Each predicted
requirement specifies a semantic field, its active value, and
one of three priority tiers. The field and value preserve
necessary contextual information, including traveler, location,
and time scope.

\begin{lstlisting}[style=judgeprompt]
{
  "intent": [
    {
      "field": "semantic_field",
      "value": "active value",
      "priority": "must_have"
    }
  ]
}
\end{lstlisting}

The following block is inserted as
\texttt{\{INTENT\_RULES\}} in both execution templates.

\begin{lstlisting}[style=judgeprompt]
Return cumulative active requirements as an intent array.
Each item has exactly field, value, and priority. Do not emit relation or scope.
- field: a concise semantic field name; use the supplied vocabulary where applicable.
  Preserve necessary context in the field or value, e.g. budget_max,
  day_2_required_activity, day_3_excluded_activity, or entity_2_mobility.
- value: the actual typed JSON value, preserving dates, travelers, limits and
  inclusion/exclusion meaning. Use numbers for numeric limits where the field
  already identifies the limit; structured values may carry necessary context.
- priority: must_have (required), preferred (tradeable preference), or optional
  (may be omitted). Infer from user requirements and trade-offs, not recency.
Preserve a tier until the user changes it; remove superseded requirements.
Split distinct requirements into separate items. Never merge items only because
they share a field name. Keep stable traveler references across turns.
Do not invent requirements, values or flight IDs absent from the available context.
Do not emit separate constraints, entities, ranked_fields or priority objects.
\end{lstlisting}

\paragraph{Interpretation of optional requirements.}
An optional requirement remains part of the active intention
until the user withdraws or supersedes it.
Its lower priority permits it to remain unsatisfied in the
selected outcome; it does not permit its omission from the
intention prediction.

\subsection{Shared Action Selection Policy}
\label{app:shared_action_prompt}

Both execution templates include a policy requiring a concrete,
evidence-supported choice even when all user requirements cannot
be jointly satisfied. The agent must retain its inferred
intention state and disclose unmet conditions separately.

The following block is inserted as
\texttt{\{BEST\_AVAILABLE\_RULES\}} in WebShop and forms the
first part of \texttt{\{TRAVEL\_SELECTION\_RULES\}} in
TravelPlanner.

\begin{lstlisting}[style=judgeprompt]
Final action policy (best available concrete choice):
- Submit your best concrete available solution even when no option satisfies all
  user conditions. Make the trade-off yourself; do not defer it to the user.
- Use your own inferred must_have/preferred/optional priorities to compare whole
  solutions. Protect must_have first, then preferred, then optional. If user
  must-have requirements conflict, choose the best achievable compromise and
  identify exactly which requirements the selected solution does not meet.
- Do not change the intention prediction to make your chosen action look compliant.
  Report unmet conditions in the action rationale; a disclosed violation is still
  a violation for evaluation.
- Select actual supported options. Do not return an unchosen list of alternatives,
  'such as', 'to be confirmed', 'select later', or 'if available' as the final choice.
- Continue researching alternatives when tools are available: a missing flight can
  require searching a ground route; an unavailable room type can require another
  room type. Do not stop at the first failed exact match.
- Never invent candidates, availability, prices or attributes. Resource restrictions
  such as minimum hotel nights and occupancy are feasibility conditions: choose a
  different supported option or adjust the stay and explicitly disclose the user
  requirement being sacrificed. Do not present an impossible booking as feasible.
- Only a genuine absence of any supported alternative after the available research
  may be reported as infeasible. This is an explicit failure, not an unresolved
  placeholder or a successful plan. A lack of a perfect match is not this exception.
\end{lstlisting}

\subsection{WebShop Execution Template}
\label{app:webshop_execution_prompt}

The WebShop prompt requests one action at a time, together with
the agent's current intention understanding. Its runtime context
contains the current user utterance, the user utterance history,
a concise trace of internal steps within the current turn, and
the current page observation.

The template below includes the intention schema in expanded
form. The two shared instruction placeholders are replaced by
the blocks in
Sections~\ref{app:shared_intent_prompt}
and~\ref{app:shared_action_prompt}, respectively.

\begin{lstlisting}[style=judgeprompt]
You are controlling a WebShop execution agent.
Decide the single best next action for the current page.
Return one JSON object only.

Allowed action_type values:
- search
- click
- buy
- back_to_search
- next_page
- prev_page

Rules:
- Use the page contents, item context, WebShop instruction text, and the trajectory user utterances only.
- A concise internal_step_trace may be provided for earlier steps in the same turn. Use it only as short within-turn memory; treat the current page as the source of truth.
- Infer the active user requirements from the latest user utterance while using earlier user utterances only as user-side trajectory context.
- If the latest user utterance changes or narrows the request, decide whether to search again based on that utterance and the returned items.
- When composing a search query, avoid negative phrasing such as "not sexy" or "not v neck". Prefer positive descriptors that imply the same intent, such as "casual modest" or "crew neck", or leave hard-to-express negatives out of the query and filter them during item inspection.
- Use action_type="buy" when the selected item page is the final rollout result. This is a virtual finalization action, not a WebShop click.
- Do not use action_type="click" with target "Buy Now" or "Buy".
- If you need to select an option on an item page, use action_type="click" with the exact option value from clickables.
- If a navigation button like "back to search", "next >", or "< prev" is the right move, prefer the dedicated action types when possible.
- When using action_type="click", the target must exactly match one available clickable string.
- When using action_type="search", include a non-empty query.
- If the current page provides too little evidence to select, inspect or navigate instead.

Required JSON schema:
{
  "current_intention_understanding": {
    "intent": [
      {
        "field": "semantic_field",
        "value": "active value",
        "priority": "must_have"
      }
    ]
  },
  "action_type": "search | click | buy | back_to_search | next_page | prev_page",
  "action_payload": {
    "query": "string when needed",
    "target": "clickable string when needed"
  },
  "rationale": "short explanation"
}

{INTENT_RULES}

{BEST_AVAILABLE_RULES}

EXECUTOR_CONTEXT_JSON:
{WEBSHOP_CONTEXT_JSON}
\end{lstlisting}

The runtime placeholder
\texttt{\{WEBSHOP\_CONTEXT\_JSON\}} is populated with a JSON
object containing the following fields. Placeholder values
below describe the supplied content rather than literal
model inputs.

\begin{lstlisting}[style=judgeprompt]
{
  "current_user_utterance": "<current user utterance>",
  "trajectory_user_utterances": [
    "<user utterances through the current turn>"
  ],
  "internal_step_trace": [
    "<serialized internal steps within this turn>"
  ],
  "page": {
    "<serialized observation fields>": "<current page evidence>"
  }
}
\end{lstlisting}

\subsection{TravelPlanner Execution Template}
\label{app:travelplanner_execution_prompt}

The TravelPlanner prompt jointly requests the current cumulative
intention and one environment action. It specifies the available
search tools, their argument formats, the notebook workflow,
and the required final itinerary submission.

The runtime context contains the user utterance history and
the public environment observation. The intention schema is
expanded below; the shared intention instructions are inserted
at \texttt{\{INTENT\_RULES\}}.

\begin{lstlisting}[style=judgeprompt]
You are the single evaluated TravelPlanner agent.
Infer the user's current cumulative intention and choose exactly one next environment action.
Return one JSON object only.

Required schema:
{
  "predicted_current_intention": {
    "intent": [
      {
        "field": "semantic_field",
        "value": "active value",
        "priority": "must_have"
      }
    ]
  },
  "action_type": "FlightSearch | AttractionSearch | AccommodationSearch | RestaurantSearch | CitySearch | GoogleDistanceMatrix | NotebookWrite | Planner",
  "action_payload": {},
  "rationale": "brief reason"
}

Action payloads:
- FlightSearch: {"origin": "city", "destination": "city", "date": "YYYY-MM-DD"}
- AttractionSearch: {"city": "city"}
- AccommodationSearch: {"city": "city"}
- RestaurantSearch: {"city": "city"}
- CitySearch: {"state": "state name"}
- GoogleDistanceMatrix: {"origin": "city", "destination": "city", "mode": "self-driving or taxi"}
- NotebookWrite: {"description": "short description of the preceding tool result"}
- Planner: {"query": "current request", "plan": {"itinerary": [...]}}

Rules:
- Use only user_utterances and public_environment_observation.
- Later utterances may add, replace, remove, relax, or reprioritize earlier requirements.
- Represent shared and person-specific requirements as separate intent items with context in their field or value. Use stable traveler IDs and natural references across turns.
- Adding or removing a traveler changes people_number; do not collapse that person's separate schedule, mobility, meal, activity, or transport need into a shared constraint.
{INTENT_RULES}
- Use real search tools, write their results to Notebook, then submit a grounded plan.
- If pending_notebook is true, the next action must be NotebookWrite.
- A Planner action must include the complete itinerary in action_payload.plan.
- Never invent tool results.

{TRAVEL_SELECTION_RULES}

CONTEXT:
{TRAVELPLANNER_CONTEXT_JSON}
\end{lstlisting}

The block \texttt{\{TRAVEL\_SELECTION\_RULES\}} consists
of the shared action-selection policy in
Section~\ref{app:shared_action_prompt}, immediately followed
by these TravelPlanner-specific instructions:

\begin{lstlisting}[style=judgeprompt]
- In the final itinerary, commit to a named accommodation for every overnight stay,
  a specific supported transport option for each required journey (including the
  trip endpoint/return), and concrete meal choices. Self-catering or packed food
  can be a concrete choice when explicitly selected, not a promise to decide later.
- Write a complete daily itinerary with day, current_city, transportation,
  breakfast, lunch, dinner, attraction and accommodation. Use '-' only when an
  item is intentionally not needed, such as lodging after the trip ends or an
  intentionally omitted attraction; explain sacrificed requirements in rationale.
- Account for accommodation per occupied/billed night, fares for all travelers,
  and meal costs for the party. Report known total costs and genuinely unpriced
  components without treating them as free or claiming a verified total budget.
\end{lstlisting}

The runtime placeholder
\texttt{\{TRAVELPLANNER\_CONTEXT\_JSON\}} contains the following
fields. The public observation is updated after environment
actions.

\begin{lstlisting}[style=judgeprompt]
{
  "user_utterances": [
    "<user utterances through the current turn>"
  ],
  "public_environment_observation": {
    "<public observation fields>": "<environment evidence>"
  }
}
\end{lstlisting}

\section{Evaluation Details and Prompts}
\label{app:evaluation}

\subsection{Evaluation Procedure}

The offline evaluation pipeline separates baseline preparation,
model-output judging, and deterministic score aggregation.
Both domains use the same high-level
\texttt{prepare--run--summarize} workflow. Each model is evaluated once per trajectory. We report mean scores over scored turns, and the Average row in Table 3 is the arithmetic mean across the eleven models.

\paragraph{Baseline preparation.}
For each turn, preparation constructs a shared baseline from the
human annotations. The baseline freezes the active gold constraints,
their priorities, atomic requirements for intention matching,
per-constraint satisfaction criteria, human feasibility, and
reference-action judgments.
The same baseline is reused across evaluated models.
Source and configuration fingerprints are recorded to prevent reuse
of incompatible cached judgments.

\paragraph{Intention judging.}
The intention judge decomposes predicted requirements into atoms
and matches them one-to-one with the frozen gold atoms.
Content correctness requires both the value and applicability
scope to match. Joint correctness additionally requires the
predicted priority to match the human annotation.
Priority is taken from the agent's original prediction and
cannot be inferred or replaced by the judge.
The judge also identifies changes relative to the previous
prediction to support change-specific evaluation.

\paragraph{Action judging.}
The action judge evaluates the submitted outcome against the frozen
criteria independently of the agent's intention prediction.
For each in-scope constraint, it returns a satisfaction verdict and
supporting evidence.
WebShop judgments use the selected product's catalog information
and actual option selections.
TravelPlanner judgments use the submitted itinerary and supporting
environment records, supplemented by deterministic cost and
accommodation-validity checks.
A claimed attribute or a stated intention alone is insufficient
evidence that an action satisfies a requirement.

\paragraph{Deterministic action scoring.}
Action evaluation uses each in-scope gold constraint field as one scoring unit. Atomization is used for intention matching and does not split a field into additional action-scoring units. For a field containing a conjunction of requirements, every component must be supported for that field to receive satisfaction credit. Scores are computed using the metrics defined in Section~\ref{sec:metrics}. Only a \texttt{satisfied} verdict receives satisfaction credit;
\texttt{violated} and \texttt{unknown} verdicts receive none.

\paragraph{WebShop selection checks.}
A selection fails these checks when there is no final product
selection, when selected options have an invalid representation,
or when a selected option name or value is not supported by the
product record.
An available option is not evidence that the agent selected it;
requirements involving size, color, or other variants are judged
using the recorded selection.
A selected ASIN that is absent from the catalog snapshot is
treated as a data error by the evaluator.

\paragraph{TravelPlanner accommodation checks.}
The accommodation checks assess minimum stays and capacity using matched accommodation records. Each itinerary day with a non-empty accommodation slot counts as one booked night. Every consecutive stay at a matched listing must satisfy that listing's minimum-night requirement. For each night with matched listings, their combined maximum occupancy must accommodate the travel party. A detected violation of either condition fails the accommodation-validity check. Minimum-stay validity is based on the booked nights, rather than
on an assumed payment for additional, unbooked nights.
These checks assess accommodation feasibility;
individual user requirements are evaluated separately
through per-constraint judgments.

\paragraph{Missing information and aggregation.}
An unknown constraint-satisfaction verdict receives no satisfaction
credit. If the human feasibility label is absent, the scorer uses
the feasible branch of $G_t$. If a human reference outcome is absent,
the scorer assumes that the reference satisfies every in-scope
constraint when computing the reference counts.
Missing gold intention is an evaluation error.

For each metric, turn-level scores are averaged over turns for
which that metric is defined. Intention and action evaluation
maintain separate coverage counts and denominators.
Evaluator, API, and data errors must be distinguished from
unsuccessful agent outcomes when interpreting this coverage.

Below, we reproduce selected instruction blocks from the
implementation and the corresponding output schemas.
Instruction wording is retained, with line wrapping adjusted for presentation. Omitted material is explicitly marked.
Placeholders denote serialized, instance-specific inputs.
The excerpts are not complete serialized API requests.

\subsection{TravelPlanner Baseline Preparation}

The baseline prompt receives the dialogue prefix, gold constraints
and priorities, the reference plan when available, and environment
records. It establishes atomic requirements and frozen criteria
that are reused across evaluated agents.

\begin{lstlisting}[style=judgeprompt]
You prepare a frozen audit baseline for ONE turn of a travel-planning dialogue.
It is shared by every evaluated agent, so it must not depend on any agent. You
see the dialogue so far, the gold constraints with tiers, the gold reference
plan (if any) and the candidate records the agents could choose from.

1. Gold atoms. Split every gold constraint into semantic atoms, one requirement
   per atom. Keep dates, day numbers, cities, travelers, negations and limits
   inside the atom. Copy the source field name into `source_field`. Never add
   information that the gold value does not state. A constraint that is already
   a single requirement becomes one atom.

[Rule 2, concerning activity requirements, omitted.]

3. Scope. Put a gold field in `out_of_scope` only if it is (a) an accessibility
   or mobility need of a traveler, or (b) the feasibility of moving around
   within a city (walking, transit, rideshare between stops). The candidate
   records cannot audit these. Nothing else is out of scope.

4. Criteria. For every in-scope gold field except budget, write `criteria`:
   one or two sentences stating exactly what a plan must do at this turn to
   satisfy the constraint, with scope resolved from the dialogue (which days,
   meals, cities, travelers) and the verbatim user quote it rests on (empty if
   the requirement comes only from the gold). Criteria restate the gold
   requirement; they never add, drop or relax it. When the dialogue conflicts
   with the gold, the criteria follow the gold and the conflict is reported as
   an annotation issue. These criteria are the only interpretation the action
   judge receives.

5. Gold plan audit. If a gold reference plan is given, judge every in-scope gold
   field except budget against it, with the action rules below: satisfied,
   violated or unknown, with a short evidence note. This establishes Gold's
   satisfied counts, NOT World Feasibility, which comes from human annotation.
   Budget, minimum nights and
   capacity of the gold plan are computed by code.

6. Annotation issues. Report conflicts between the gold constraints and the
   dialogue: a constraint the user withdrew, a value contradicting the latest
   utterance, a missing constraint the user clearly stated, an empty candidate
   pool. Describe them; do not correct the gold.
\end{lstlisting}

The builder additionally appends the action-audit rules, any
configured few-shot examples, the output schema, field-coverage
instructions, and the turn-specific payload.
The required output schema is:

\begin{lstlisting}[style=judgeprompt]
{
  "gold_atoms": [{
    "atom_id": "g1",
    "source_field": "exact gold field",
    "value": "one requirement"
  }],
  "activity_requirements": [{
    "kind": "include | exclude | order | limit | free_time",
    "date": "YYYY-MM-DD or null",
    "requirement": "...",
    "source_turn": 0,
    "quote": "..."
  }],
  "out_of_scope": [{
    "field": "exact gold field",
    "reason": "..."
  }],
  "constraint_criteria": [{
    "gold_field": "exact gold field",
    "criteria": "what a plan must do at this turn",
    "quote": "verbatim user words, or empty"
  }],
  "gold_plan_judgments": [{
    "gold_field": "exact gold field",
    "status": "satisfied | violated | unknown",
    "evidence": "..."
  }],
  "annotation_issues": ["..."]
}
\end{lstlisting}

\subsection{WebShop Baseline Preparation}

WebShop uses the current human intention snapshot as the authority
for scoring requirements. Dialogue may clarify a requirement but
cannot introduce or remove scoring fields. The following block is
appended after the WebShop evidence rules.

\begin{lstlisting}[style=judgeprompt]
Prepare a shared baseline, independent of tested models.
Cover every active constraint exactly once in criteria; split gold_atoms only for intention matching.
The sole authority for active constraint fields, values and priorities is INPUT.gold, normalized from gold_current_intention.
Use exactly the fields in INPUT.gold.constraints.
Dialogue, delta annotations, product facts and agent claims must never add, remove, replace or relax these constraints.
Dialogue only clarifies their meaning.
If another source disagrees, report an annotation issue and retain the current Gold snapshot.
Judge every constraint against Gold selection if present, otherwise return no Gold judgments.
Never judge World Feasibility or alter human tiers.
Report annotation conflicts without repairing them.
Return JSON matching:
{
  "gold_atoms": [{
    "atom_id": "g1",
    "source_field": "exact field",
    "value": "one requirement"
  }],
  "constraint_criteria": [{
    "gold_field": "exact field",
    "criteria": "all required evidence",
    "quote": "verbatim user quote or empty"
  }],
  "gold_plan_judgments": [{
    "gold_field": "exact field",
    "status": "satisfied | violated | unknown",
    "evidence": "product field and supporting text"
  }],
  "annotation_issues": []
}

INPUT:
{INPUT_JSON}
\end{lstlisting}

\subsection{Shared Intention Judge}

Both domains use the same intention-matching instructions.
The judge separates semantic content, applicability scope,
and changes relative to the agent's previous prediction.
Priority correctness is subsequently computed from the original
predicted tiers rather than supplied by the judge. 

\begin{lstlisting}[style=judgeprompt]
You compare ONE agent's predicted intent at one turn with the frozen gold
atoms. You do not see any itinerary.

1. Split every predicted item into atoms with the same rule as the gold atoms:
   one requirement per atom, keeping dates, cities, travelers, negations and
   limits; never add information the item does not state. Give each atom the
   index of the predicted item it came from.

2. Match predicted atoms to gold atoms one-to-one: each gold atom and each
   predicted atom is used at most once. If two predicted atoms express the same
   gold atom, match the better one and leave the other unmatched. A renamed but
   clearly equivalent field may match.

3. value_match is true only if the atom's value is semantically correct for the
   gold atom's current value: Day 2 does not satisfy Day 1, a budget ceiling is
   not an exact spending target, an exclusion is not an inclusion.
   Also return scope_match as an explicit boolean. It is true only when the
   entity, traveler, city, day/time and applicability scope match the Gold atom.
   Same value for a different entity/scope is not a correct constraint.
   For unscoped requirements with no conflicting scope, return true when the
   Gold atom is matched; unmatched predictions must have scope_match=false.
   Never borrow a Gold entity/scope absent from the prediction unless the
   reference is unambiguously resolved by the supplied dialogue.

4. change_vs_previous compares each atom with the agent's own previous-turn
   prediction: unchanged if a previous item expresses the same requirement with
   the same value (renaming or rephrasing is not a change), changed if it
   expresses the same requirement with a different value, new if absent before.
   At the first turn every atom is new.

5. priority_change_vs_previous compares the priority of the same predicted
   requirement (including its entity/scope) to the agent's previous prediction:
   unchanged when the tier is identical, changed when it differs, new when no
   previous matching requirement exists. A value change alone does not imply a
   priority change. At the first turn every atom is new. Missing priority stays
   missing; never infer or fill in a tier from dialogue, Gold or importance.
   Match content without using priority to choose the Gold atom. Priority
   correctness is calculated in code from the original predicted item.

Return exactly one JSON object:
{
  "pred_atoms": [{
    "source_index": 0,
    "field": "...",
    "value": "one requirement",
    "gold_atom_id": "g1 or null",
    "value_match": true,
    "scope_match": true,
    "change_vs_previous": "new | changed | unchanged",
    "priority_change_vs_previous": "new | changed | unchanged"
  }]
}

Every predicted item index must yield at least one atom.

PAYLOAD:
{INPUT_JSON}
\end{lstlisting}

\paragraph{Schema notation.}
In the displayed intention-judge schema, expressions such as
``g1 or null'' describe alternative output values rather than
literal strings to be returned.
A matched prediction uses a gold atom identifier such as
\texttt{"g1"}; an unmatched prediction uses JSON \texttt{null},
without quotation marks.

\subsection{TravelPlanner Action Judge}

The action judge evaluates the final plan against the baseline's
frozen criteria. It uses environment records as evidence and
separates constraint satisfaction from the agent's disclosure of
unmet requirements.

\begin{lstlisting}[style=judgeprompt]
You audit ONE turn of ONE agent's travel plan against a frozen baseline. You do
not see the agent's intent prediction; judge only the plan.

1. Final plan. Start from the itinerary as written. Apply only explicit final
   revisions stated in the action (for example "replace X with Y", "switch the
   Day 2 hotel to Z"); record each in `applied_revisions` with its quote. Judge
   the revised plan as one whole: anything that conflicts with the final
   revision is judged on the final version. Mark every final_plan slot a
   revision changed with "revised": true.

2. Record mapping. For every non-empty transportation, meal and accommodation
   slot of the final plan, give `record_name`: the exact restaurant or listing
   name, or the flight number, from the candidate records; null when nothing
   matches. Code pre-matched names are supplied; keep them unless a revision
   changes the slot or the pre-match is clearly wrong. A slot may name several
   listings (a group split across rooms); join their exact names with " + ".

3. Facts come from the candidate records. Take prices, ratings, room types,
   house rules, minimum nights, maximum occupancy, cuisines and flight times
   from the records, never from the agent's wording or rationale. An entity
   absent from the records cannot satisfy a constraint that depends on its
   attributes: that constraint is violated, not unknown.

4. Criteria are the requirement. Each field to judge carries frozen criteria
   from the shared baseline; judge the plan against them. You do not see the
   dialogue, and must not reinterpret, add, drop or relax requirements. Per-city
   values such as {"Denver": 4} apply only to that city.

[Rules 5--9, concerning activities, party size, hotels,
house rules, and meals, omitted.]

10. Budget is priced by code; do not judge it.

11. unknown is a last resort: only when the records genuinely lack the
    attribute that decides the constraint.

12. Disclosures. In `unmet_constraint_disclosures`, list every gold field that
    the action explicitly tells the user is not met (naming the requirement or
    its value), with the quote. Vague trade-off language does not count.
    Disclosure never changes action_status.

13. Annotation issues. Report data problems that prevent a fair audit, such as
    an empty candidate pool.

[Configured few-shot examples and output schema omitted.]

Judge every field listed in fields_to_judge exactly once, against its frozen criteria.

PAYLOAD:
{INPUT_JSON}
\end{lstlisting}

The structured response includes the resolved final plan,
explicit revisions, per-constraint verdicts and evidence,
disclosures of unmet requirements and annotation issues.
Budget computation and accommodation-validity checks are handled separately in code. The complete TravelPlanner judging instructions, including
Rules 5--9 omitted from the excerpt, are provided in
\path{src/eval/rules/travelplanner_v2.md}.
These rules specify the treatment of activities, party size,
accommodation attributes, house rules, and meals.

\subsection{WebShop Action Judge}

WebShop requires positive evidence for requested product properties and distinguishes available options from options actually selected. Selected evidence rules are reproduced below. In this excerpt, ``actual Gold'' refers to the human reference
selection, ``World Feasibility'' refers to the human-provided
feasibility label, and the ``Must count gate'' refers to $G_t$
defined in Section~\ref{sec:metrics}.
These terms are retained from the implementation.

\begin{lstlisting}[style=judgeprompt]
These are per-constraint evidence rules, identical for Agent and actual Gold.
The judge does not compute scores, priorities or World Feasibility.

1. A product without evidence supporting a requested constraint FAILS that
   constraint. Return violated with a missing_evidence explanation, not a
   speculative satisfied. Legacy unknown verdicts also receive zero credit.
   Agent claims, common knowledge and plausible product properties are not evidence.

[Rule 2, concerning the missing-reference scoring convention, omitted.]

3. Judge the selected ASIN AND selected options. An available size/color is not
   a selected size/color. Do not invent a default when the environment does not
   record it. Inspect full descriptions, bullets and attributes, not only titles.

4. Each frozen constraint is one scoring unit. For AND requirements every
   component needs support; for OR requirements one alternative suffices.
   Do not split, merge, relax or remove fields. Report ambiguous booleans,
   duplicate requirements and stale annotations as annotation issues at baseline.

5. Black does not prove matte black; metal does not prove steel. Hooks do not
   prove hook-and-loop closure. An included outdoor antenna does not prove an
   included indoor antenna. Compatibility does not prove a component is included.
   A product explicitly labeled glam fails an exclusion of glam even if it is
   also described as modern. Never invent attribute values or units.

6. A negative ingredient requirement needs supporting evidence of absence;
   lack of a keyword alone is not proof. Follow the baseline's interpretation
   of ambiguous fields such as artificial_ingredients=true.

7. Use the selected product/variant price and the criterion's exact comparator
   (< versus <=); distinguish per-item prices, pack contents and quantity.

8. Disclosures are exact quotes from the selected action rationale, reported
   separately. They do not change whether a constraint is satisfied. Python
   applies the human feasible/not-feasible Must count gate and soft 2:1 ratio.

Audit only this selected action against frozen criteria.
Do not see/infer the agent intention or use Gold performance to change verdicts.

Return JSON:
{
  "constraint_judgments": [{
    "gold_field": "exact field",
    "action_status": "satisfied | violated | unknown",
    "evidence": "specific product evidence or missing fact"
  }],
  "unmet_constraint_disclosures": [{
    "gold_field": "field",
    "quote": "exact rationale quote"
  }]
}

{INPUT_JSON}
\end{lstlisting}

\section{Additional Error Analysis}
\label{app:error_analysis}

Table~\ref{tab:failure_by_intention_recovery} reports constraint violations and invalid bookings in failed turns, conditioned on whether the agent fully recovered the user's intention.

\begin{table}[H]
\centering
\small
\setlength{\tabcolsep}{3pt}
\renewcommand{\arraystretch}{1.15}

\begin{tabularx}{\columnwidth}{
    @{}>{\raggedright\arraybackslash}Xrr@{}
}
\toprule
\textbf{Violation category}
& \shortstack{\textbf{Full intention}\\\textbf{recovery}}
& \shortstack{\textbf{Incomplete}\\\textbf{recovery}} \\
\midrule
\multicolumn{3}{@{}l}{\textit{Travel Planner}} \\
Budget                & 79.4\% & 39.6\% \\
Preferences           & 10.3\% & 39.7\% \\
Itinerary structure   &  3.8\% & 29.5\% \\
Activity              &  0.0\% &  1.3\% \\
Invalid booking (min.\ stay / capacity) & 33.1\% & 50.4\% \\
\midrule
\multicolumn{3}{@{}l}{\textit{WebShop}} \\
Product attributes    & 60.4\% & 70.4\% \\
Specification options & 41.7\% & 21.4\% \\
Product category      & 28.6\% & 26.5\% \\
Price                 &  7.1\% & 15.0\% \\
\bottomrule
\end{tabularx}

\vspace{3pt}
\begin{minipage}{0.96\columnwidth}
\footnotesize
\textit{Note.} Percentages indicate the fraction of failed turns exhibiting each violation category. Invalid booking denotes a violation of accommodation minimum-stay or capacity rules. A single failed turn may violate multiple constraints, so categories are not mutually exclusive and percentages within a column need not sum to 100\%. For TravelPlanner, itinerary structure includes transportation, check-in, and scheduling constraints; preferences include room type, ratings, dining, and policy constraints. For WebShop, specification options include size,
color, and quantity.
\end{minipage}

\caption{Constraint violations and invalid bookings in action failures, conditioned on intention recovery.}
\label{tab:failure_by_intention_recovery}
\end{table}

\section{Sensitivity to Soft-Constraint Weights}
\label{app:weight_sensitivity}

Action Score assigns weights of 2 and 1 to satisfied preferred and optional constraints, respectively. We recompute scores from the same per-turn satisfaction judgments under six alternatives: linear weightings of 1:1, 3:1, and 5:1; a lexicographic scheme prioritizing preferred constraints; the minimum of the preferred and optional satisfaction ratios relative to the human reference; and the squared 2:1 ratio. The Hard Success gate is unchanged.

Table~\ref{tab:weight_sensitivity} compares each resulting ranking with the original 2:1 ranking across all eleven models. Kendall's $\tau$ is at least 0.964 in both domains, no model moves by more than one position, and the top three models and their ordering remain unchanged. Average Action Score changes by at most 0.5 percentage points under linear weightings and 3.5 percentage points under the nonlinear variants. These results indicate that the ranking conclusions are stable across the tested scoring schemes.

\begin{table}[t]
\centering
\small
\setlength{\tabcolsep}{4pt}
\begin{tabular}{@{}lcc@{}}
\toprule
Scoring scheme & TravelPlanner & WebShop \\
 & $\tau$ / $\Delta r_{\max}$ 
 & $\tau$ / $\Delta r_{\max}$ \\
\midrule
Linear 1:1       & 1.000 / 0 & 1.000 / 0 \\
Linear 3:1       & 1.000 / 0 & 1.000 / 0 \\
Linear 5:1       & 0.964 / 1 & 1.000 / 0 \\
Lexicographic    & 0.964 / 1 & 1.000 / 0 \\
Min.\ tier ratio & 0.964 / 1 & 0.964 / 1 \\
Squared 2:1 ratio & 0.964 / 1 & 1.000 / 0 \\
\bottomrule
\end{tabular}
\caption{Sensitivity of model rankings to alternative
soft-constraint scoring schemes.
Kendall's $\tau$ measures agreement with the original
2:1 ranking; $\Delta r_{\max}$ denotes the maximum
absolute rank change across eleven models.}
\label{tab:weight_sensitivity}
\end{table}

\section{Prediction Precision and Obsolete Retention}
\label{app:precision_retention}

Intention Recovery does not penalize additional predictions.
We therefore provide two supplementary diagnostics.
\textit{Precision} measures the fraction of predicted constraint
atoms matched to active gold constraints.
\textit{Obsolete Retention} measures how often a previously
correctly recorded requirement remains immediately after the user
overrides, relaxes, or removes it, without matching any active
constraint. Retention is identified through exact matches to the
agent's previous prediction and may miss paraphrased cases.

Precision should be interpreted cautiously because unmatched
predictions may include valid fixed task parameters omitted from
the gold intention or differences in constraint granularity.
Obsolete Retention more specifically measures failures to discard
superseded requirements.
\begin{table}
\centering
\small
\begin{tabular}{lcc}
\toprule
Domain & Precision $\uparrow$ & Obsolete retention $\downarrow$ \\
\midrule
TravelPlanner & 54.4 & 14.0 \\
WebShop       & 84.5 & 13.8 \\
\bottomrule
\end{tabular}
\caption{Supplementary intention diagnostics (\%),
averaged across eleven models.}
\label{tab:precision_retention}
\end{table}

\end{document}